\documentclass[conference]{IEEEtran}
\usepackage{times}

\usepackage[numbers]{natbib}
\usepackage{multicol}
\usepackage[bookmarks=true]{hyperref}
\usepackage{graphicx}

\usepackage{multirow}
\usepackage{multicol}
\usepackage{graphicx}
\usepackage{xspace}
\usepackage{xcolor}
\usepackage{caption}
\usepackage{wrapfig}
\usepackage{bbding}
\usepackage{pifont}

\usepackage{booktabs}
\usepackage{float}
\usepackage{amsmath}
\usepackage{amssymb}
\usepackage{bm}

\usepackage{dsfont}

\usepackage{hyperref}
\usepackage[capitalise, nameinlink]{cleveref}

\usepackage{colortbl}
\definecolor{ourcolor}{HTML}{99e0eb}
\definecolor{ourblue}{HTML}{27a2c3}

\definecolor{tablecolor}{HTML}{ccf2f5}

\definecolor{tablecolor2}{HTML}{ffcdb4}
\definecolor{grey}{rgb}{0.9, 0.9, 0.9}
\usepackage{amssymb}

\usepackage{listings}
\definecolor{gred}{rgb}{0.859,0.267,0.216}
\definecolor{ggreen}{rgb}{0.059,0.616,0.345}

\definecolor{deepblue}{HTML}{27a2c3}

\definecolor{deepred}{HTML}{B71C1C}

\usepackage[font=small,labelfont=bf]{caption}

\usepackage[font=footnotesize,labelfont=bf]{caption}

\definecolor{lblue}{HTML}{ffb114}
\definecolor{ogreen}{HTML}{2E7D32}
\definecolor{bred}{HTML}{BF360C}
\definecolor{newbrown}{HTML}{795548}

\hypersetup{
    colorlinks=true,
    linkcolor=deepblue,
    urlcolor=deepblue,
    citecolor=deepblue,
    filecolor=brown,
}

\usepackage{multicol}
\usepackage{multirow}
\usepackage{colortbl}
\usepackage{booktabs}
\usepackage{bbding}
\usepackage{graphicx}
\let\labelindent\relax
\usepackage{enumitem}

\usepackage{bbding}

\usepackage{pifont}
\usepackage{xfrac}
\usepackage{arydshln}

\begin{document}


\title{\textit{Now You See That:} Learning End-to-End \\Humanoid Locomotion from Raw Pixels}

\author{
    \authorblockN{
        Wandong Sun\textsuperscript{1,2} \quad
        Yongbo Su\textsuperscript{1,2} \quad
        Leoric Huang\textsuperscript{2} \quad
        Alex Zhang\textsuperscript{2} \quad
        Dwyane Wei\textsuperscript{2} \quad
        Mu San\textsuperscript{2} \quad 
        \\
        Daniel Tian\textsuperscript{2} \quad 
        Ellie Cao\textsuperscript{2} \quad 
        Baoshi Cao\textsuperscript{1} \quad
        Yang Liu\textsuperscript{1} \quad
        Finn Yan\textsuperscript{2} \quad 
        Ethan Xie\textsuperscript{2} \quad
        Zongwu Xie\textsuperscript{1} \quad
    }
    \authorblockA{
        \textsuperscript{1}Harbin Institute of Technology \quad
        \textsuperscript{2}HONOR Robotics Team\quad
        \\
        \vspace{5pt}
        \href{https://github.com/Hellod035/Now_You_See_That}{\texttt{\nolinkurl{https://github.com/Hellod035/Now_You_See_That}}}
        \vspace{-5pt}
        }
}


\twocolumn[{%
            \renewcommand\twocolumn[1][]{#1}%
            \maketitle
            \vspace{-0.45cm}
            \begin{center}
                \centering
                \captionsetup{type=figure}
                \includegraphics[width=1.0\textwidth]{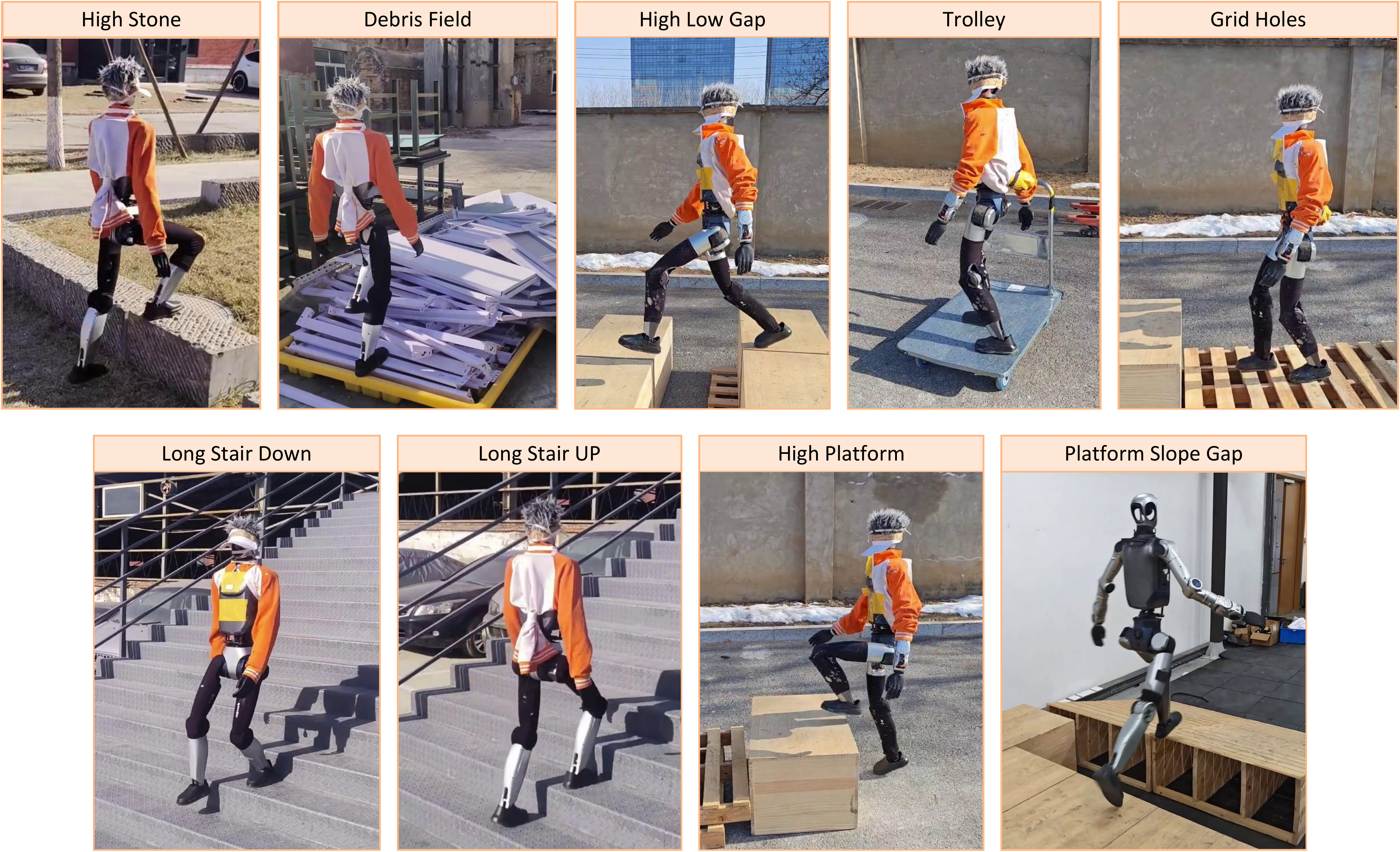}
                \vspace{-0.17in}
                \caption{\textbf{Overview.} Our end-to-end vision-based humanoid locomotion policy enables robust traversal across diverse challenging terrains, including high stones, long staircases (both ascending and descending), debris fields, gaps with varying heights, trolleys, high platforms, grid holes, and platform-slope-gap combinations. All behaviors emerge from a single unified policy trained with raw depth images.}%
                \label{fig:teaser}
            \end{center}
            \vspace{0.04in}
        }]

\begin{abstract}
    Achieving robust vision-based humanoid locomotion remains challenging due to two fundamental issues: the sim-to-real gap introduces significant perception noise that degrades performance on fine-grained tasks, and training a unified policy across diverse terrains is hindered by conflicting learning objectives.
    To address these challenges, we present an end-to-end framework for vision-driven humanoid locomotion.
    For robust sim-to-real transfer, we develop a high-fidelity depth sensor simulation that captures stereo matching artifacts and calibration uncertainties inherent in real-world sensing. We further propose a vision-aware behavior distillation approach that combines latent space alignment with noise-invariant auxiliary tasks, enabling effective knowledge transfer from privileged height maps to noisy depth observations.
    For versatile terrain adaptation, we introduce terrain-specific reward shaping integrated with multi-critic and multi-discriminator learning, where dedicated networks capture the distinct dynamics and motion priors of each terrain type.
    We validate our approach on two humanoid platforms equipped with different stereo depth cameras. The resulting policy demonstrates robust performance across diverse environments, seamlessly handling extreme challenges such as high platforms and wide gaps, as well as fine-grained tasks including bidirectional long-term staircase traversal.
\end{abstract}
\IEEEpeerreviewmaketitle{}

\begin{table*}[t]
    \centering
    \small
    \vspace{5pt}
    \resizebox{\textwidth}{!}{
        \begin{tabular}{l c c c c c c}
            \toprule
            \textbf{Method}                                                     & \textbf{Representation} & \textbf{End-to-End} & \textbf{Noise Modeling} & \textbf{Long-Term Deploy} & \textbf{Fine Locomotion} & \textbf{Extreme Parkour} \\
            \midrule
            Long et al.~\cite{long2024learning}                                 & Elevation Map           & \ding{55}           & Moderate                & \ding{55}                 & \ding{51}                & \ding{55}                \\
            Sun et al.~\cite{sun2025learning}                                   & Elevation Map           & \ding{55}           & Moderate                & \ding{55}                 & \ding{51}                & \ding{55}                \\
            He et al.~\cite{he2025attention}                                    & Elevation Map           & \ding{55}           & Moderate                & \ding{55}                 & \ding{51}                & \ding{55}                \\
            Ben et al.~\cite{ben2025gallantvoxelgridbasedhumanoid}              & Voxel                   & \ding{55}           & Moderate                & \ding{51}                 & \ding{51}                & \ding{55}                \\
            Zhuang et al.~\cite{zhuang2024humanoid}                             & End-to-End Vision       & \ding{51}           & Moderate                & \ding{51}                 & \ding{55}                & \ding{51}                \\
            Song et al.~\cite{song2025gaitadaptiveperceptivehumanoidlocomotion} & Vision-to-Elevation     & \ding{55}           & Moderate                & \ding{51}                 & \ding{51}                & \ding{55}                \\
            \textbf{Ours}                                                       & End-to-End Vision       & \ding{51}           & Comprehensive           & \ding{51}                 & \ding{51}                & \ding{51}                \\
            \bottomrule
        \end{tabular}
    }
    \caption{Comparison of perceptive humanoid locomotion methods. Representation indicates the terrain perception approach. Noise Modeling indicates the comprehensiveness of depth sensor simulation. Long-Term Deploy indicates drift-free operation capability. Fine Locomotion indicates support for precise movements like stair climbing. Extreme Parkour indicates support for dynamic maneuvers across challenging obstacles.}    \label{tab:related_work_comparison}
    \vspace{-15pt}
\end{table*}

\section{Introduction}\label{sec:introduction}

Vision-based humanoid locomotion provides a compelling benchmark for embodied
intelligence, requiring robots to traverse diverse terrains, from extreme
obstacles like high platforms and wide gaps to fine-grained challenges such as
continuous staircases, using onboard visual and proprioceptive sensing. Unlike
quadrupedal locomotion where static stability tolerates moderate control
errors~\cite{hwangbo2019learning}, humanoid robots operate in inherently
unstable regimes~\cite{Goswami2004RateOC, McGeer01041990} that demand precise
coordination between perception and action. Vision-based humanoid locomotion
faces two fundamental challenges: \textbf{(1)} perception noise from the
sim-to-real gap severely degrades performance on fine-grained tasks requiring
centimeter-level accuracy, and \textbf{(2)} training a single policy to handle
diverse terrain scenarios remains difficult due to conflicting learning
objectives across heterogeneous environments.

Current perception approaches for legged robots can be broadly categorized into
LiDAR-based and vision-based methods. Elevation map-based methods fuse LiDAR
height measurements with odometry to construct terrain
representations~\cite{Fankhauser2014RobotCentricElevationMapping,
    Fankhauser2018ProbabilisticTerrainMapping, miki2022elevation, kalman1960new},
achieving success on both quadrupedal~\cite{Miki_2022, hoeller2024anymal,
    lee2020learning} and humanoid platforms~\cite{long2025learning,
    sun2025learning, li2024reinforcement}. Voxel-based 3D occupancy
grids~\cite{ben2025gallantvoxelgridbasedhumanoid,
    nahrendra2024obstacleawarequadrupedallocomotionresilient, hornung13auro}
explicitly represent volumetric geometry. However, both LiDAR-based approaches
suffer from limited sensing frequency and inherent latency, constraining their
applicability for highly dynamic maneuvers. Vision-based methods using depth
cameras~\cite{agarwal2023legged, cheng2024extreme, zhuang2023robot,
    yang2021learning, loquercio2023learning} offer higher bandwidth ego-centric
perception, yet their application to humanoid platforms remains limited due to
sim-to-real gaps from imperfect depth sensor modeling.


Beyond perception, learning unified control across heterogeneous terrains poses
additional challenges. Standard reinforcement learning with a single value
function struggles to capture these diverse reward landscapes, leading prior
work to train separate specialist policies before
distillation~\cite{hoeller2024anymal, zhuang2023robotparkourlearning},
incurring substantial training overhead.

In this work, we present an end-to-end vision-driven locomotion framework
addressing both challenges. For robust perception transfer, we introduce
comprehensive depth augmentation operators that reproduce realistic stereo
camera imperfections, combined with vision-aware behavior distillation that
aligns latent spaces through noise-invariant auxiliary tasks. For unified
terrain mastery, we employ terrain-specific reward shaping with multi-critic
reinforcement learning~\cite{mysore2022multi} and
multi-discriminator\cite{Hardy_2019} adversarial motion priors, where dedicated
networks capture distinct dynamics and terrain-appropriate behaviors. We
validate our approach on two humanoid platforms with different depth cameras,
demonstrating robust performance across both extreme challenges (high
platforms, wide gaps) and fine-grained locomotion tasks (long term staircases
in both ascending and descending directions). Our contributions are:

\begin{itemize}
    \item \textbf{Realistic depth sensor simulation:} A comprehensive augmentation pipeline simulating stereo matching artifacts, depth-dependent noise, optical distortions, and calibration uncertainties for seamless sim-to-real transfer.

    \item \textbf{Vision-aware behavior distillation:} A distillation framework combining latent space shaping with noise-invariant auxiliary tasks, transferring locomotion knowledge from privileged observations to noisy depth inputs.

    \item \textbf{Multi-critic and Multi-discriminator terrain learning:} Terrain-specific reward shaping with dedicated value networks and discriminators capturing distinct dynamics and motion priors across diverse terrains within a single unified policy.

    \item \textbf{Cross-platform validation:} Extensive real-world experiments on two humanoid robots with different stereo depth cameras, demonstrating successful traversal across both extreme and fine-grained terrains in diverse indoor and outdoor environments.
\end{itemize}

\section{Related Work}\label{sec:related_work}

\subsection{Perceptive Legged Locomotion}

Early work on legged locomotion primarily focused on height-based terrain
perception using LiDAR sensors combined with odometry
systems~\cite{Fankhauser2014RobotCentricElevationMapping,
    Fankhauser2018ProbabilisticTerrainMapping}. These methods construct elevation
maps that enable quadrupedal robots to traverse challenging terrains by
providing dense geometric information for foothold
planning~\cite{lee2020learning, hoeller2024anymal, miki2022elevation}. Recent
humanoid locomotion systems have adopted similar
approaches~\cite{radosavovic2024real, li2024reinforcement, long2025learning,
    he2025attentionbasedmapencodinglearning}. However, the dynamic stability of
bipedal locomotion significantly amplifies odometry drift issues, cumulative
errors in map registration directly cause foothold misalignment and falls,
particularly in high-impact scenarios such as running, jumping, or stair
navigation.

To eliminate odometry dependency, depth camera-based approaches have emerged as
an alternative. Direct depth-to-action learning has demonstrated robust terrain
traversal in quadrupedal systems~\cite{agarwal2023legged, cheng2024extreme,
    zhuang2023robot, zhuang2023robotparkourlearning, yang2021learning,
    loquercio2023learning}, where static stability provides inherent tolerance to
perception noise. However, extending these methods to humanoid platforms
remains challenging~\cite{sun2025dpl, sun2025learning}.
An alternative approach reconstructs elevation maps directly from egocentric
depth observations without external
odometry~\cite{song2025gaitadaptiveperceptivehumanoidlocomotion}, avoiding
cumulative drift but introducing additional complexity compared to end-to-end
approaches.


\subsection{Data Augmentation in Visual Reinforcement Learning}

Data augmentation (DA) has become a fundamental technique in visual
reinforcement learning for improving both sample efficiency and generalization
ability~\cite{rad, drq, drq_v2, 2022image_da_survey, 2022comprehensive_survey}.
Early work explored various image transformations, with random cropping
emerging as particularly effective for sample-efficient training~\cite{drq,
    drq_v2}. Beyond simple transformations, spectrum-based augmentation~\cite{SRM}
and saliency-guided approaches~\cite{SGQN} have been proposed to improve
generalization. However, existing domain randomization techniques primarily
focus on simple perturbations~\cite{dr1, dr2, chen2021understanding}, lacking
systematic modeling of structured sensor artifacts present in real-world depth
cameras. For depth-based perception, comprehensive augmentation strategies must
simulate realistic sensor imperfections, including stereo fusion artifacts,
depth-dependent noise patterns, optical distortions, and calibration
variations~\cite{svea}, combined with appropriate preprocessing techniques to
handle sensor-specific characteristics.

Recent advances have shown that auxiliary training objectives are essential for
improving generalization beyond visual augmentation
alone~\cite{auxiliary_tasks_in_rl}. Auxiliary task approaches have proven
essential~\cite{auxiliary_tasks_in_rl}: contrastive learning methods like
CURL~\cite{curl} maximize mutual information between augmented views to learn
invariant representations~\cite{DRIML, huang2021towards}, while consistency
regularization techniques such as DrAC~\cite{drac} explicitly penalize policy
differences across augmentations to enforce noise-invariant features.
Pre-trained visual representations have also shown promise for improving sample
efficiency and generalization~\cite{RRL, PVR, R3M, mvp, pieg, hansen2022pre}.
Masked autoencoder-based approaches~\cite{mae, mlr} and predictive
representation learning~\cite{Predictive_1, Predictive_2} offer alternative
strategies for learning robust features.
\begin{figure}[t]
    \centering
    \includegraphics[width=0.45\textwidth]{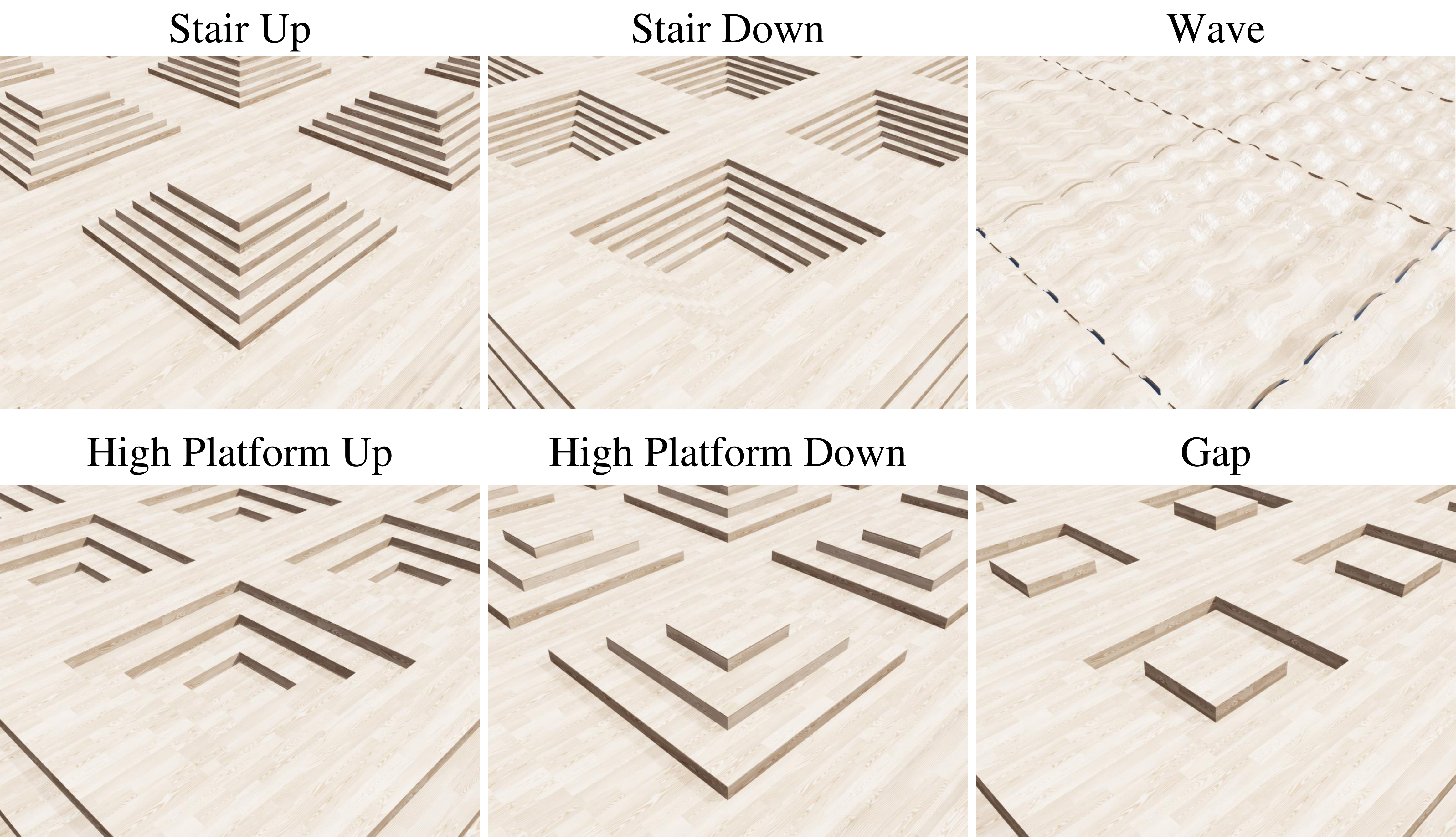}
    \caption{Diverse terrain types used during training. Each terrain type contains 20 difficulty levels for curriculum learning.}
    \label{fig:terrains}
    \vspace{-15pt}
\end{figure}

\begin{table*}[ht]
    \centering
    \small
    \begin{tabular}{cllll}
        \toprule
        \textbf{Order} & \textbf{Operation}  & \textbf{Schedule} & \textbf{Parameters}                                                            & \textbf{Description}        \\
        \midrule
        \multicolumn{5}{l}{\textit{Preparation Stage}}                                                                                                                          \\
        \cdashline{1-5}\noalign{\vskip 0.3mm}
        1              & Camera intrinsics   & Once at startup   & $s_{\text{h}}, s_{\text{v}} \sim \mathcal{U}(0.90, 1.10)$                      & Focal length variations     \\
        2              & Camera extrinsics   & Once at startup   & $\Delta \mathbf{p} \sim \mathcal{U}(-0.05, 0.05)^3$ m                          & Mounting position offset    \\
                       &                     &                   & $\Delta \boldsymbol{\theta} \sim \mathcal{U}(-0.10, 0.10)^3$ rad               & Mounting orientation offset \\
        3              & Observation delay   & Once at startup   & $d_{\text{frame}} \sim \mathcal{U}[2, 4]$ frames                               & Processing pipeline latency \\
        \midrule
        \multicolumn{5}{l}{\textit{Processing Pipeline}}                                                                                                                        \\
        \cdashline{1-5}\noalign{\vskip 0.3mm}
        1              & Stereo fusion       & Per frame         & $\tau \sim \mathcal{U}[0.05, 0.20]$                                            & Disparity consistency check \\
        2              & Random convolution  & Per frame         & $w_{i,j,k,l} \sim \mathcal{U}(-0.05, 0.05)$                                    & Optical aberrations         \\
        3              & Gaussian noise      & Per frame         & $c_0, c_1, c_2 \sim \mathcal{U}(-0.03, 0.03)$                                  & Distance-dependent noise    \\
        4              & Perlin noise        & Per frame         & $c_0^{\text{p}}, c_1^{\text{p}}, c_2^{\text{p}} \sim \mathcal{U}(-0.02, 0.02)$ & Time-dependent noise        \\
        5              & Scale randomization & Per frame         & $s_i \sim \mathcal{U}(0.90, 1.10)$                                             & Calibration errors          \\
        6              & Pixel failures      & Per frame         & $p_{\text{zero}} = p_{\text{max}} = 0.001$                                     & Dead/saturated pixels       \\
        7              & Depth clipping      & Per frame         & $[d_{\min}, d_{\max}] = [0.3, 2.0]$ m                                          & Valid sensing range         \\
        8              & Spatial cropping    & Per frame         & $(t, b, l, r) = (3, 3, 4, 4)$ pixels                                           & Edge distortion removal     \\
        \bottomrule
    \end{tabular}
    \caption{Vision Augmentation Pipeline: Operations, Schedule, Parameters and Descriptions}
    \label{tab:augmentation_pipeline}
    \vspace{-10pt}
\end{table*}

\begin{figure*}[htbp]
    \centering
    \includegraphics[width=0.9\textwidth]{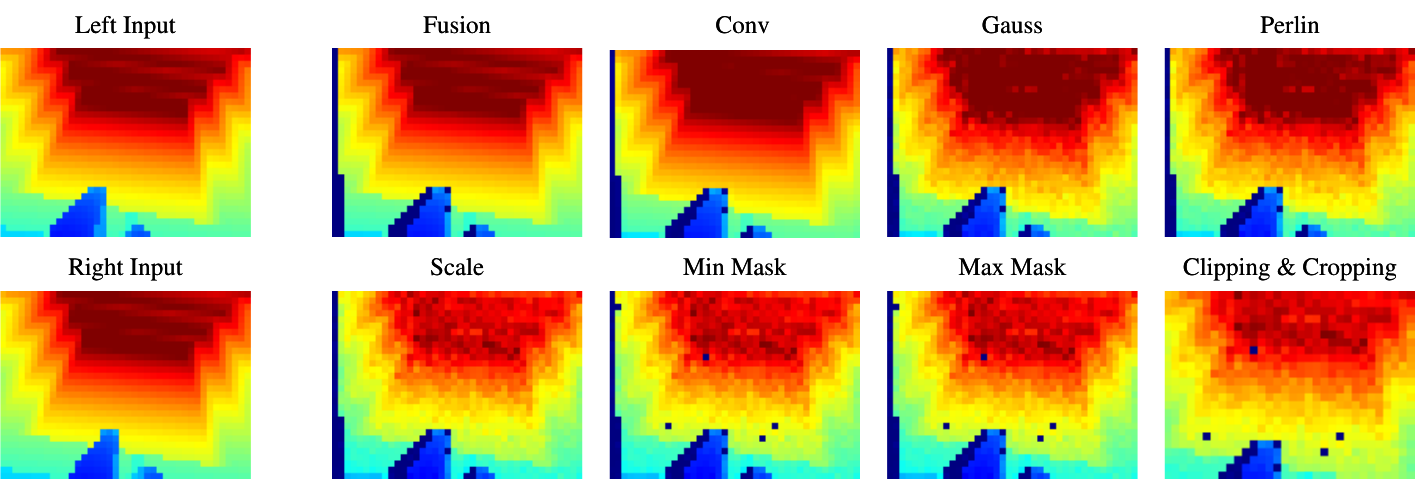}
    \caption{Visualization of the depth augmentation pipeline. Starting from clean left and right depth images, the pipeline sequentially applies: (1) stereo fusion, (2) random convolution, (3) Gaussian noise, (4) Perlin noise, (5) scale randomization, (6) zero pixel failures, (7) max pixel failures, (8) depth clipping and spatial cropping to produce realistic depth observations for sim-to-real transfer.}
    \label{fig:augmentation_fig}
    \vspace{-10pt}
\end{figure*}

\section{Method}\label{sec:method}

\begin{figure}[t]
    \centering
    \includegraphics[width=0.45\textwidth]{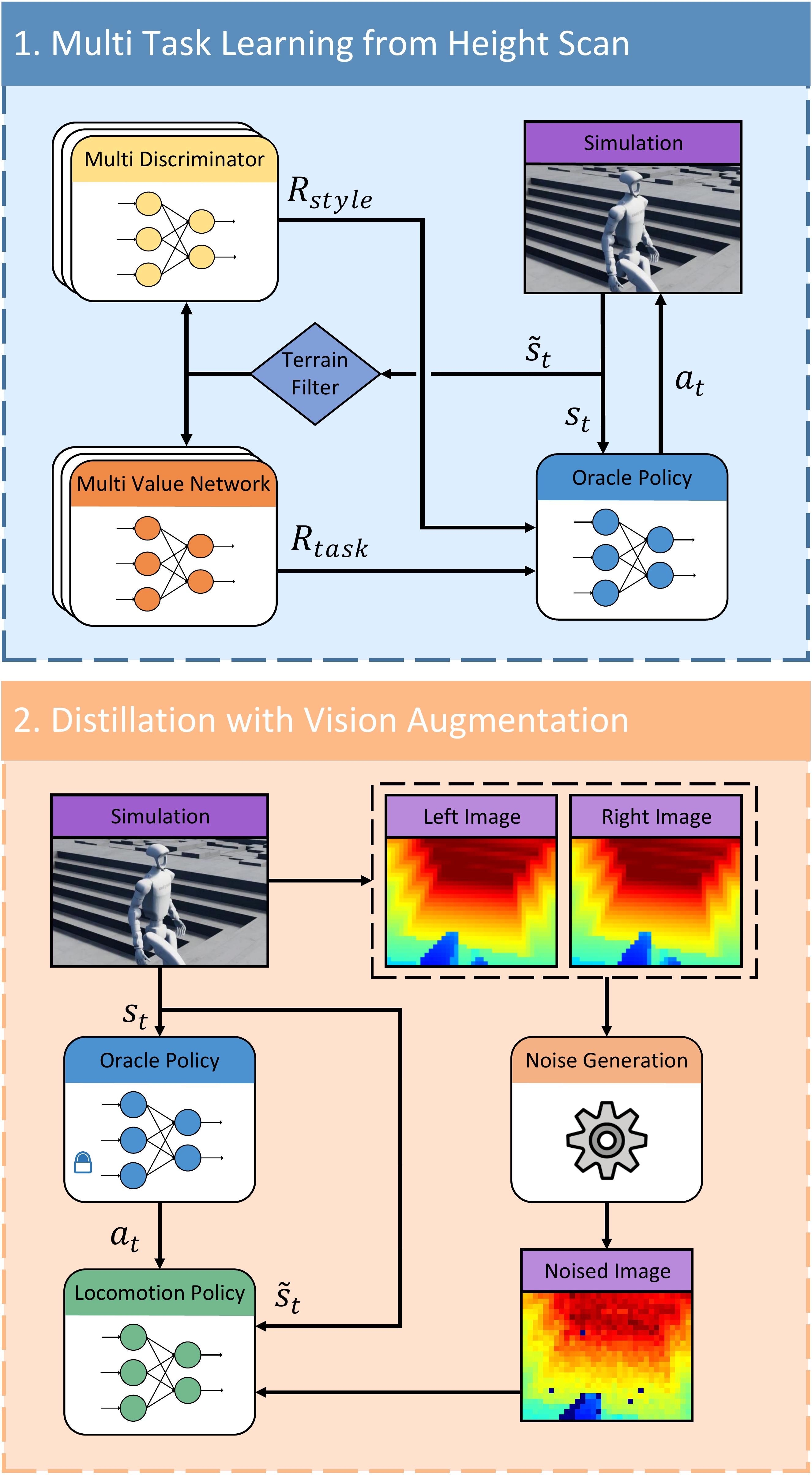}
    \caption{\textbf{Method Overview.} Our framework consists of two stages: (1) \textit{Privileged RL Training}: A teacher policy is trained with height scan observations using multi-critic and multi-discriminator learning, where terrain-specific reward shaping and dedicated value networks handle diverse terrain categories (stairs/platforms, gaps, rough terrain). (2) \textit{Vision-Aware Distillation}: The privileged policy is distilled into a deployment policy operating on augmented depth images, combining behavior cloning with denoising objectives for robust sim-to-real transfer.}
    \label{fig:method_overview}
    \vspace{-15pt}
\end{figure}

We present an end-to-end framework for vision-based humanoid locomotion that
addresses both perception transfer and unified terrain control. As illustrated
in Figure~\ref{fig:method_overview}, our approach follows a two-stage training
pipeline. In the first stage, we train a privileged policy using height scan
observations with multi-critic and multi-discriminator reinforcement learning,
where terrain-specific reward shaping and dedicated value networks enable
efficient learning across diverse scenarios. In the second stage, we distill
the privileged policy into a deployment policy that operates directly on depth
images, using vision-aware behavior distillation with comprehensive depth
augmentation to ensure robust sim-to-real transfer.

\subsection{Realistic Depth Sensor Simulation}\label{sec:vision_augmentation}

Real depth cameras exhibit structured imperfections that standard simulation
pipelines fail to capture. We introduce a comprehensive augmentation pipeline
that sequentially applies eight operators to simulate realistic sensor
artifacts, bridging the sim-to-real gap for robust policy transfer.

\subsubsection{Stereo Depth Fusion}

Stereo cameras reconstruct depth through binocular correspondence matching,
producing characteristic hole patterns in occluded and textureless regions. We
simulate this process by rendering images from left and right viewpoints with
baseline distance $b$. For each pixel $(u, v)$ in the left image, we compute
disparity-based correspondence:
\begin{equation}
    u_r = u - \frac{f_x b}{d_{\text{left}}(u, v) + \epsilon}
\end{equation}
where $f_x$ is the focal length and $\epsilon = 10^{-6}$ prevents division by zero. The fused depth applies a consistency check:
\begin{equation}
    d_{\text{fused}}(u, v) = \begin{cases}
        d_{\text{left}}(u, v) & \text{if } |d_{\text{left}} - d_{\text{right}}(u_r, v)| < \tau \cdot d_{\text{left}} \\
        0                     & \text{otherwise}
    \end{cases}
\end{equation}
where $\tau \in [0.05, 0.20]$ controls matching sensitivity. Pixels failing the consistency check are marked invalid, reproducing the hole patterns caused by occlusions and texture-poor surfaces.

\subsubsection{Depth-Dependent Noise}

Real depth measurements exhibit noise profiles that grow nonlinearly with
distance due to the inverse relationship between disparity and depth. We model
this effect with quadratic scaling:
\begin{equation}
    \tilde{d}(u, v) = d(u, v) + \mathcal{N}(0, \sigma^2(d)), \quad \sigma(d) = |c_0 + c_1 d + c_2 d^2|
\end{equation}
where $c_0, c_1, c_2 \sim \mathcal{U}(-0.03, 0.03)$ are randomly sampled per environment to capture inter-device variations. The absolute value ensures non-negative noise amplitude while preserving the randomization of noise characteristics across different simulated sensors.

\subsubsection{Structured Noise Patterns}

Environmental interference and sensor artifacts manifest as spatially
correlated noise. We generate multi-octave Perlin noise to simulate these
structured patterns:
\begin{equation}
    n_{\text{perlin}}(u, v) = \sum_{o=0}^{4} 0.5^o \cdot \mathcal{P}(2^o u, 2^o v)
\end{equation}
where $\mathcal{P}(\cdot, \cdot)$ denotes the standard Perlin noise function that generates smooth, continuous pseudo-random values through gradient interpolation. We use 5 octaves with persistence 0.5 to balance between large-scale intensity variations and fine-grained texture, matching the multi-scale nature of real sensor interference patterns.
The noise is applied with depth-dependent amplitude: $\tilde{d} = d + \sigma_{\text{p}}(d) \cdot n_{\text{perlin}}$, where $\sigma_{\text{p}}(d) = |c_0^{\text{p}} + c_1^{\text{p}} d + c_2^{\text{p}} d^2|$ follows the same quadratic scaling with absolute value to ensure non-negative amplitude.

\subsubsection{Optical Distortions}

Lens aberrations introduce local geometric distortions that vary across the
image plane. We model these effects through randomized convolution:
\begin{equation}
    \tilde{d} = d * (W + I)
\end{equation}
where $W \in \mathbb{R}^{3 \times 3}$ with $w_{ij} \sim \mathcal{U}(-0.05, 0.05)$ and $I$ is the identity kernel, ensuring the distortion remains centered around the original measurement.

\subsubsection{Calibration Uncertainties}

Manufacturing variations and imperfect calibration cause systematic measurement
errors. We randomize depth scaling $s \sim \mathcal{U}(0.90, 1.10)$ to simulate
factory calibration drift, perturb camera intrinsics through aperture scaling,
and add extrinsic offsets to camera position and orientation. Additionally, we
model pixel failures through random masking with probability $p_{\text{zero}} =
    p_{\text{max}} = 0.001$.

\subsubsection{Preprocessing}

We clip depth values to the valid sensing range $[0.3, 2.0]$ m, normalize to
$[0, 1]$, and crop from $30 \times 40$ to $24 \times 32$ pixels to reduce
peripheral distortion artifacts. Observation delay sampled from $\mathcal{U}[2,
        4]$ frames simulates processing pipeline latency.
Table~\ref{tab:augmentation_pipeline} summarizes the complete augmentation
pipeline with parameter ranges.

\subsection{Privileged Reinforcement Learning}\label{sec:multi_critic_rl}

Parkour terrains exhibit fundamentally different dynamics, and a single value
function struggles to capture these diverse reward landscapes. We address this
challenge through terrain-specific reward shaping with dedicated critic and
discriminator networks.

\subsubsection{Height Scan Observations}

The privileged policy receives dense geometric information via ray casting over
a 1.6 m $\times$ 1.0 m ego-centric window with 0.05 m resolution, yielding a
grid of $21 \times 33 = 693$ height samples. This terrain representation is
subsequently transferred to depth-based perception through distillation. The
architecture of the depth encoder is detailed in
Appendix~\ref{append:network_architectures}.

\subsubsection{Terrain-Specific Reward Shaping}

We partition terrains into $K=3$ categories with specialized reward components:
(1) \textit{stairs and platforms}, encompassing stair ascending, stair
descending, platform ascending, and platform descending; (2) \textit{gap
    crossing}; and (3) \textit{rough terrain} locomotion.
Detailed reward formulations are provided in
Appendix~\ref{append:terrain_spec_rewards}.

\subsubsection{Multi-Critic and Multi-Discriminator Architecture}

We maintain $K=3$ critic networks $\{V_k\}_{k=1}^K$ and $K=3$ discriminator
networks $\{D_k\}_{k=1}^K$, each specialized for a terrain category: (1) stairs
and platforms, (2) gap crossing, and (3) rough terrain. The critics share a
common backbone while maintaining separate output heads, enabling efficient
parameter sharing while preserving terrain-specific value estimation. During
training, the appropriate critic and discriminator are selected based on
terrain labels:
\begin{equation}
    V(s_t) = V_{k(s_t)}(s_t), \quad D(s_t) = D_{k(s_t)}(s_t)
\end{equation}
where $k(s_t) \in \{1, 2, 3\}$ identifies the current terrain category. Each
critic learns the value landscape for its specific dynamics, while the actor
receives policy gradients from the terrain-appropriate critic. Similarly, each
discriminator provides style rewards tailored to the motion characteristics
required for its terrain type.

\subsubsection{Motion Priors}

We incorporate Adversarial Motion Priors~\cite{Peng_2021} to guide natural
locomotion. The discriminator operates on torso-centric observations:
\begin{equation}
    \Phi_{\text{amp}} = [\mathbf{q}_{\text{rel}}, \dot{\mathbf{q}}_{\text{rel}}, \mathbf{v}_{\text{torso}}^b, \omega_{\text{torso}}^b, \mathbf{g}^b, \mathbf{p}_{\text{body}}^b, \mathbf{q}_{\text{body}}^b]
\end{equation}
We collect separate motion datasets for each terrain category to provide appropriate style guidance.

\subsubsection{Advanced Domain Randomization}

We apply comprehensive randomization to robot dynamics to ensure robust
sim-to-real transfer. Table~\ref{tab:robot_randomization} details the complete
randomization schedule covering contact properties, body parameters, actuator
dynamics, and external disturbances.

\begin{table}[t]
    \centering

    \small
    \begin{tabular}{ll}
        \toprule
        \textbf{Parameter}        & \textbf{Range}         \\
        \midrule
        \multicolumn{2}{l}{\textit{Contact Properties}}    \\
        \cdashline{1-2}\noalign{\vskip 0.3mm}
        Static/Dynamic friction   & $(0.4, 1.2)$           \\
        Restitution               & $(0.0, 0.4)$           \\
        \midrule
        \multicolumn{2}{l}{\textit{Body Properties}}       \\
        \cdashline{1-2}\noalign{\vskip 0.3mm}
        Mass scaling              & $(0.8, 1.2)$           \\
        COM offset                & $\pm 0.03$ m           \\
        Default joint offset      & $\pm 0.03$ rad         \\
        \midrule
        \multicolumn{2}{l}{\textit{Actuator Dynamics}}     \\
        \cdashline{1-2}\noalign{\vskip 0.3mm}
        Armature scaling          & $(0.5, 1.5)$           \\
        Stiffness/Damping scaling & $(0.9, 1.1)$           \\
        \midrule
        \multicolumn{2}{l}{\textit{External Disturbances}} \\
        \cdashline{1-2}\noalign{\vskip 0.3mm}
        Push velocity             & $\pm 1.2$ m/s          \\
        IMU bias                  & $\pm 0.04$ rad/s       \\
        \bottomrule
    \end{tabular}
    \caption{Robot Dynamics Domain Randomization}
    \label{tab:robot_randomization}
    \vspace{-15pt}
\end{table}

\begin{table*}[t]
    \centering
    \resizebox{\textwidth}{!}{
        \begin{tabular}{l|ccc|ccc|ccc|ccc|ccc}
            \toprule
                                & \multicolumn{3}{c|}{\textbf{Stairs Up}} & \multicolumn{3}{c|}{\textbf{Stairs Down}} & \multicolumn{3}{c|}{\textbf{Gaps}} & \multicolumn{3}{c|}{\textbf{Platform}} & \multicolumn{3}{c}{\textbf{Average}}                                                                                                                                                                                                                                                                                                                       \\
            \textbf{Method}     & SR                                      & P$\downarrow$                             & PDR$\downarrow$                    & SR                                     & P$\downarrow$                        & PDR$\downarrow$              & SR                           & P$\downarrow$                & PDR$\downarrow$              & SR                           & P$\downarrow$                & PDR$\downarrow$              & SR                           & P$\downarrow$                & PDR$\downarrow$              \\
            \midrule
            No Augmentation     & 45.3$_{\pm 1.8}$                        & 52.3$_{\pm 1.2}$                          & 68.4$_{\pm 2.1}$                   & 42.1$_{\pm 2.0}$                       & 48.7$_{\pm 1.1}$                     & 71.2$_{\pm 2.3}$             & 44.8$_{\pm 1.7}$             & 51.2$_{\pm 1.0}$             & 69.5$_{\pm 1.9}$             & 39.7$_{\pm 2.2}$             & 54.6$_{\pm 1.3}$             & 74.3$_{\pm 2.5}$             & 43.0$_{\pm 1.9}$             & 51.7$_{\pm 1.2}$             & 70.9$_{\pm 2.2}$             \\
            Partial Aug.        & 61.2$_{\pm 1.5}$                        & 43.8$_{\pm 0.9}$                          & 42.6$_{\pm 1.4}$                   & 57.4$_{\pm 1.6}$                       & 41.2$_{\pm 0.8}$                     & 45.3$_{\pm 1.5}$             & 59.8$_{\pm 1.4}$             & 42.5$_{\pm 0.7}$             & 43.8$_{\pm 1.3}$             & 53.6$_{\pm 1.8}$             & 46.1$_{\pm 1.0}$             & 48.7$_{\pm 1.6}$             & 58.0$_{\pm 1.6}$             & 43.4$_{\pm 0.9}$             & 45.1$_{\pm 1.5}$             \\
            Standard DR         & 65.4$_{\pm 1.3}$                        & 41.5$_{\pm 0.8}$                          & 38.2$_{\pm 1.2}$                   & 62.1$_{\pm 1.4}$                       & 39.4$_{\pm 0.7}$                     & 40.6$_{\pm 1.3}$             & 63.7$_{\pm 1.2}$             & 40.2$_{\pm 0.6}$             & 39.1$_{\pm 1.1}$             & 56.8$_{\pm 1.6}$             & 44.3$_{\pm 0.9}$             & 45.2$_{\pm 1.4}$             & 62.0$_{\pm 1.4}$             & 41.4$_{\pm 0.8}$             & 40.8$_{\pm 1.3}$             \\
            Humanoid Parkour    & 74.2$_{\pm 1.1}$                        & 38.2$_{\pm 0.7}$                          & 28.5$_{\pm 1.0}$                   & 71.5$_{\pm 1.2}$                       & 36.5$_{\pm 0.6}$                     & 30.2$_{\pm 1.1}$             & 72.8$_{\pm 1.0}$             & 37.1$_{\pm 0.5}$             & 29.4$_{\pm 0.9}$             & 65.5$_{\pm 1.4}$             & 40.8$_{\pm 0.8}$             & 35.6$_{\pm 1.2}$             & 71.0$_{\pm 1.2}$             & 38.2$_{\pm 0.7}$             & 30.9$_{\pm 1.1}$             \\
            \midrule
            Direct RL           & 57.3$_{\pm 1.9}$                        & 48.6$_{\pm 1.1}$                          & 55.2$_{\pm 1.8}$                   & 53.8$_{\pm 2.1}$                       & 45.8$_{\pm 1.0}$                     & 58.4$_{\pm 2.0}$             & 55.6$_{\pm 1.8}$             & 47.2$_{\pm 0.9}$             & 56.8$_{\pm 1.7}$             & 49.3$_{\pm 2.3}$             & 51.4$_{\pm 1.2}$             & 62.1$_{\pm 2.2}$             & 54.0$_{\pm 2.0}$             & 48.3$_{\pm 1.1}$             & 58.1$_{\pm 1.9}$             \\
            Single Critic/Disc. & 85.6$_{\pm 0.8}$                        & 34.6$_{\pm 0.5}$                          & 18.3$_{\pm 0.6}$                   & 82.3$_{\pm 0.9}$                       & 32.8$_{\pm 0.4}$                     & 20.1$_{\pm 0.7}$             & 83.8$_{\pm 0.7}$             & 33.5$_{\pm 0.4}$             & 19.2$_{\pm 0.5}$             & 76.3$_{\pm 1.1}$             & 37.8$_{\pm 0.6}$             & 25.4$_{\pm 0.8}$             & 82.0$_{\pm 0.9}$             & 34.7$_{\pm 0.5}$             & 20.8$_{\pm 0.7}$             \\
            BC Only             & \underline{89.2}$_{\pm 0.7}$            & \underline{32.4}$_{\pm 0.4}$              & \underline{15.6}$_{\pm 0.5}$       & \underline{86.5}$_{\pm 0.8}$           & \underline{30.6}$_{\pm 0.4}$         & \underline{17.2}$_{\pm 0.6}$ & \underline{87.8}$_{\pm 0.6}$ & \underline{31.2}$_{\pm 0.3}$ & \underline{16.3}$_{\pm 0.4}$ & \underline{80.5}$_{\pm 0.9}$ & \underline{35.2}$_{\pm 0.5}$ & \underline{21.8}$_{\pm 0.7}$ & \underline{86.0}$_{\pm 0.8}$ & \underline{32.4}$_{\pm 0.4}$ & \underline{17.7}$_{\pm 0.6}$ \\
            \midrule
            \textbf{Ours}       & \textbf{99.2}$_{\pm 0.3}$               & \textbf{28.5}$_{\pm 0.3}$                 & \textbf{5.2}$_{\pm 0.2}$           & \textbf{98.6}$_{\pm 0.4}$              & \textbf{27.2}$_{\pm 0.3}$            & \textbf{6.1}$_{\pm 0.3}$     & \textbf{99.3}$_{\pm 0.2}$    & \textbf{25.8}$_{\pm 0.2}$    & \textbf{4.5}$_{\pm 0.2}$     & \textbf{98.4}$_{\pm 0.5}$    & \textbf{29.4}$_{\pm 0.4}$    & \textbf{7.2}$_{\pm 0.3}$     & \textbf{98.9}$_{\pm 0.4}$    & \textbf{27.7}$_{\pm 0.3}$    & \textbf{5.8}$_{\pm 0.3}$     \\
            \bottomrule
        \end{tabular}
    }
    \caption{Performance on RDT-Bench across four terrain configurations. SR: Success Rate (\%), P: Average Power ($\times 10^{1}$ W), PDR: Power Degradation Ratio (\%). All methods are evaluated under CycleGAN-augmented realistic depth noise. Results report mean$_{\pm \text{std}}$ over 5 random seeds. Best results are in \textbf{bold}, second best are \underline{underlined}. $\downarrow$ indicates lower is better.}
    \label{tab:main_results}
    \vspace{-15pt}
\end{table*}

\subsection{Vision-Aware Behavior Distillation}\label{sec:distillation}

We transfer the privileged policy to a deployment policy operating on depth
images through DAgger-style distillation. The student policy interacts with the
environment while the teacher provides supervision, enabling learning from its
own state distribution.

\subsubsection{Behavior Cloning}

The deployment policy minimizes the action discrepancy with the privileged
policy:
\begin{equation}
    \mathcal{L}_{\text{behavior}} = \mathbb{E}_{s_t}\left[\left\|\mu_{\text{deploy}}(s_t) - \mu_{\text{priv}}(s_t)\right\|_2^2\right]
\end{equation}
where the student receives augmented depth images with proprioceptive observations, while the teacher receives clean height scans.

\subsubsection{Denoising Objective}

To ensure robust feature extraction under sensor noise, we enforce consistency
between clean and augmented depth representations:
\begin{equation}
    \mathcal{L}_{\text{denoise}} = \mathbb{E}_{d, \tilde{d}}\left[\left\|E(d) - E(\tilde{d})\right\|_2^2\right]
\end{equation}
where $E(\cdot)$ is the depth encoder, $d$ is clean depth, and $\tilde{d}$ is its augmented counterpart.

\subsubsection{Feature Regularization}

We regularize the depth encoder features to avoid collapse by matching the
\emph{batch-wise feature distribution} to a standard normal prior. Let $z =
    E(d) \in \mathbb{R}^{N \times D}$ denote the encoder outputs.
We estimate a diagonal Gaussian $q(z) = \mathcal{N}(\boldsymbol{\mu},
    \mathrm{diag}(\boldsymbol{\sigma}^2))$ using empirical statistics:
\begin{equation}
    \mu_j = \frac{1}{N} \sum_{i=1}^{N} z_{i,j}, \quad
    \sigma_j^2 = \frac{1}{N} \sum_{i=1}^{N} (z_{i,j} - \mu_j)^2 + \epsilon
\end{equation}
where $\epsilon$ is a small constant for numerical stability. We then minimize
the KL divergence between this estimated distribution and the standard normal
prior:
\begin{equation}
    \mathcal{L}_{\text{kl}}
    = \mathrm{KL}\!\left(\mathcal{N}(\boldsymbol{\mu}, \mathrm{diag}(\boldsymbol{\sigma}^2)) \,\|\, \mathcal{N}(\mathbf{0}, \mathbf{I})\right)
\end{equation}

\subsubsection{Combined Objective}

The total distillation loss balances all components:
\begin{equation}
    \mathcal{L}_{\text{total}} = \mathcal{L}_{\text{behavior}} + \lambda_{\text{denoise}} \mathcal{L}_{\text{denoise}} + \lambda_{\text{kl}} \mathcal{L}_{\text{kl}}
\end{equation}
where $\lambda_{\text{denoise}} = \lambda_{\text{kl}} = 0.1$. The depth encoder output replaces the height scan embedding, enabling vision-based deployment.

\begin{figure*}[htbp]
    \centering
    \includegraphics[width=1.0\textwidth]{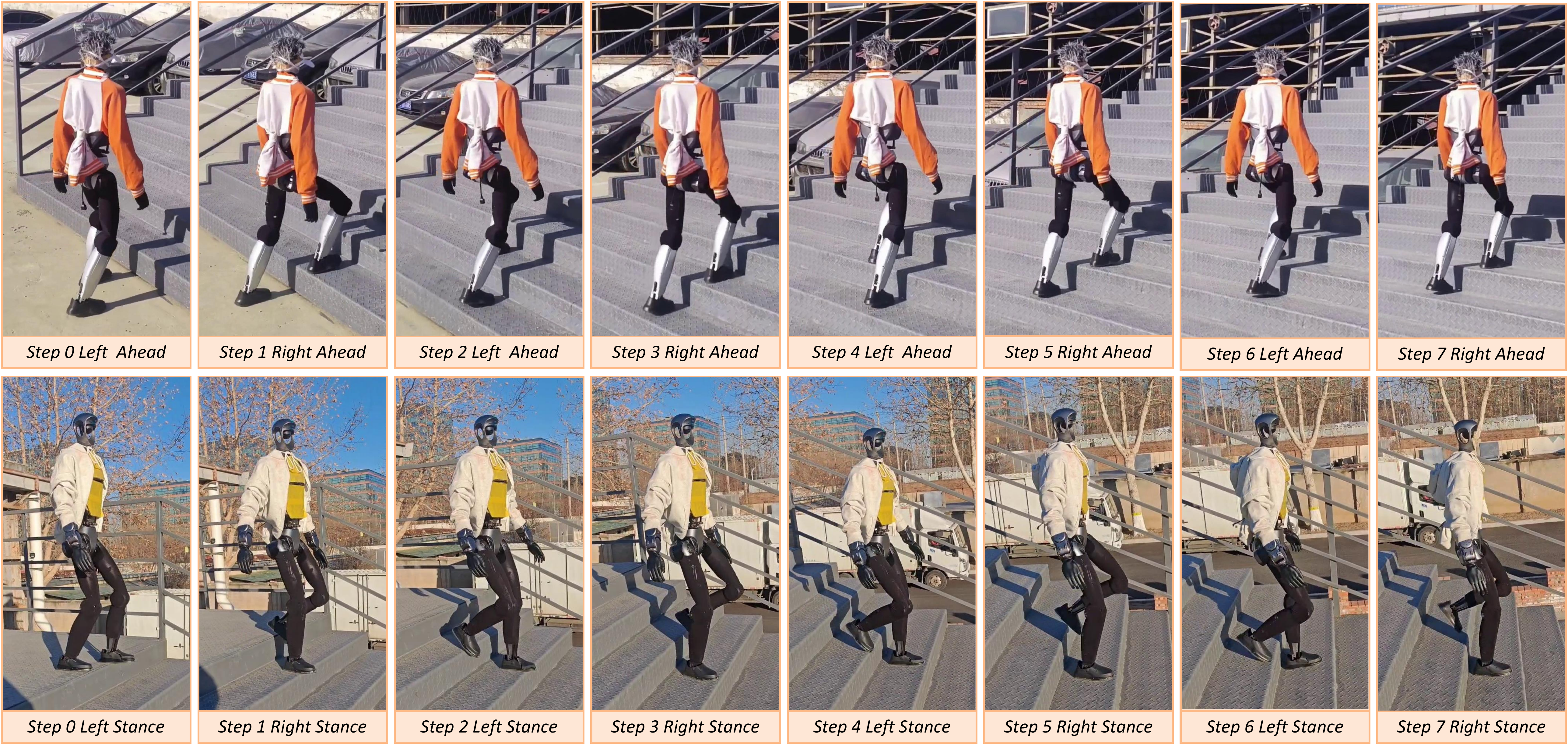}
    \caption{Real-world deployment sequences demonstrating stair traversal. Top row: ascending stairs with anticipatory leg lifting. Bottom row: descending stairs with controlled foot placement. The policy executes smooth gait patterns without any real-world fine-tuning.}
    \label{fig:up_down_stair}
    \vspace{-15pt}
\end{figure*}

\section{Experiments}\label{sec:experiments}

We evaluate our approach through comprehensive simulation benchmarks and
real-world deployment on a custom humanoid robot equipped with an Orbbec Gemini
336L depth camera. Our experiments address the following questions: (1) Does
realistic depth augmentation improve sim-to-real transfer compared to standard
domain randomization? (2) How does multi-critic learning contribute to
performance across diverse terrains? (3) Do the distillation objectives improve
policy robustness?

\subsection{Experimental Setup}\label{sec:exp_setup}

\subsubsection{Real-World Depth Transfer Benchmark}

Standard simulation benchmarks fail to capture the perception challenges of
real-world deployment. We construct the Real-World Depth Transfer Benchmark
(RDT-Bench) by training a CycleGAN~\cite{CycleGAN2017} model on unpaired depth
observations from simulation and real-world deployment. During evaluation,
policies receive CycleGAN-augmented depth images while operating in the physics
simulator, enabling large-scale quantitative comparison under realistic
perception conditions. Importantly, CycleGAN is used \textit{exclusively for
    evaluation}, each method employs its own augmentation strategy during training,
ensuring fair comparison that reveals true sim-to-real transfer capability.

\subsubsection{Evaluation Terrains}

We evaluate on four representative terrain configurations spanning three
categories:
\begin{itemize}
    \item \textbf{Stairs:} 15 cm step height and 30 cm step depth, both ascending and descending
    \item \textbf{Gap:} 0.45 m width requiring accurate depth perception for crossing
    \item \textbf{Platform:} 0.40 m height requiring precise stepping up and down
\end{itemize}
Each configuration is tested across 1024 parallel environments for 100
episodes, providing statistically robust performance estimates.

\subsubsection{Baselines}

We compare against the following methods:

\begin{itemize}
    \item \textbf{Humanoid Parkour Learning}~\cite{zhuang2024humanoid}: State-of-the-art vision-based humanoid locomotion using standard domain randomization with Gaussian noise and random dropout.
    \item \textbf{No Augmentation}: Depth images with only simple Gaussian noise ($\sigma=0.01$) and random pixel dropout ($p=0.001$).
    \item \textbf{Partial Augmentation}: Depth-dependent noise and preprocessing only, without stereo fusion, structured noise, or optical distortion.
    \item \textbf{Standard DR}: Domain randomization following~\cite{zhuang2024humanoid} with hole filling and random masking.
    \item \textbf{Single Critic/Disc.}: Our method with a single shared critic and discriminator for all terrains, instead of terrain-specific ones.
    \item \textbf{Direct RL}: End-to-end reinforcement learning directly from depth images without privileged distillation.
    \item \textbf{BC Only}: Behavior cloning with only $\mathcal{L}_{\text{behavior}}$, removing denoising and KL regularization losses.
\end{itemize}

\subsubsection{Metrics}

We evaluate methods using three complementary metrics:

\begin{itemize}
    \item \textbf{Success Rate (SR):} Percentage of episodes where the robot traverses the terrain without falling.
    \item \textbf{Average Power (P):} Average mechanical power consumption computed as $P = \frac{1}{T} \sum_t \|\tau_t \odot \dot{q}_t\|_2$, where $\tau_t$ denotes joint torques and $\dot{q}_t$ denotes joint velocities. Lower values indicate more efficient locomotion.
    \item \textbf{Power Degradation Ratio (PDR):} Relative power increase under realistic noise, defined as $\text{PDR} = (P_{\text{RDT}} - P_{\text{clean}}) / P_{\text{clean}} \times 100\%$. Lower PDR indicates greater robustness to sensor artifacts.
\end{itemize}

\subsection{Main Results}\label{sec:main_results}

Table~\ref{tab:main_results} presents comprehensive results on RDT-Bench. Our
method achieves 98.9\% average success rate with the lowest power consumption
(27.7 $\times 10^{1}$ W) and minimal power degradation (5.8\% PDR),
substantially outperforming all baselines across every terrain and metric.

\textbf{Perception Transfer.} The comparison between augmentation strategies
reveals the critical importance of realistic depth simulation. No Augmentation
achieves only 43.0\% success with severe power degradation (70.9\% PDR),
demonstrating that policies trained with simple Gaussian noise fail
catastrophically under realistic sensor artifacts. Our full augmentation
pipeline achieves 98.9\% success with only 5.8\% PDR, indicating consistent,
efficient control despite perception noise.

\textbf{Comparison with Prior Work.} Humanoid Parkour
Learning~\cite{zhuang2024humanoid} achieves 71.0\% success with 30.9\% PDR. Our
method outperforms this baseline by 27.9\% in success rate while reducing PDR
by 25.1\%, demonstrating that our approach not only improves task completion
but also maintains power-efficient locomotion under realistic perception
conditions.

\textbf{Architecture Contributions.} Single Critic/Disc. achieves 82.0\% success
with 20.8\% PDR, while our multi-critic and multi-discriminator approach
achieves 98.9\% success with 5.8\% PDR.
The terrain-specific value functions
and discriminators enable more stable policy optimization by learning distinct
motion priors for each terrain type.
Direct RL without privileged distillation shows the
highest PDR (58.1\%), confirming that the two-stage training pipeline is
essential for efficient vision-based control.

\subsection{Ablation Studies}\label{sec:ablation}

\subsubsection{Augmentation Pipeline Analysis}

Table~\ref{tab:ablation_augmentation} analyzes the contribution of each
augmentation component. Stereo Fusion provides the largest improvement by
simulating characteristic hole patterns that cause significant perception
failures on real hardware. Depth-Dependent Noise and Calibration Uncertainties
offer moderate gains by modeling distance-varying uncertainty and device
variations. Even components with smaller SR impact contribute to control
smoothness, validating the importance of comprehensive sensor modeling.

\begin{table}[h]
    \centering
    \resizebox{0.48\textwidth}{!}{
        \begin{tabular}{lccc}
            \toprule
            \textbf{Configuration}        & \textbf{SR (\%)}             & \textbf{P ($\times 10^{1}$ W)}$\downarrow$ & \textbf{PDR (\%)}$\downarrow$ \\
            \midrule
            Full Pipeline                 & \textbf{98.9}$_{\pm 0.4}$    & \textbf{27.7}$_{\pm 0.3}$                  & \textbf{5.8}$_{\pm 0.2}$      \\
            \midrule
            w/o Stereo Fusion             & 90.4$_{\pm 0.6}$             & 34.1$_{\pm 0.5}$                           & 17.5$_{\pm 0.4}$              \\
            w/o Depth-Dependent Noise     & 92.1$_{\pm 0.5}$             & 32.5$_{\pm 0.4}$                           & 14.9$_{\pm 0.3}$              \\
            w/o Structured Noise (Perlin) & \underline{96.2}$_{\pm 0.3}$ & \underline{29.4}$_{\pm 0.3}$               & \underline{8.3}$_{\pm 0.2}$   \\
            w/o Optical Distortions       & 96.9$_{\pm 0.3}$             & 28.7$_{\pm 0.3}$                           & 7.2$_{\pm 0.2}$               \\
            w/o Calibration Uncertainties & 94.5$_{\pm 0.4}$             & 31.0$_{\pm 0.4}$                           & 11.7$_{\pm 0.3}$              \\
            w/o Preprocessing             & 95.6$_{\pm 0.4}$             & 30.1$_{\pm 0.3}$                           & 9.9$_{\pm 0.3}$               \\
            \bottomrule
        \end{tabular}
    }
    \caption{Ablation study on depth augmentation components. Each row removes one component from the full pipeline. Results report mean$_{\pm \text{std}}$ over 5 random seeds.}
    \label{tab:ablation_augmentation}
    \vspace{-5pt}
\end{table}

\subsubsection{Multi-Critic Architecture}

Table~\ref{tab:ablation_critic} compares single-critic/discriminator and
multi-critic/discriminator architectures. The multi-critic/discriminator
approach provides consistent improvements across all terrain categories in both
success rate (+15.1\% to +22.5\%) and power degradation (-13.5\% to -18.2\%
PDR), indicating that terrain-specific value functions and discriminators
enable more stable gradient estimation during training.

\begin{table}[h]
    \centering
    \resizebox{0.48\textwidth}{!}{
        \begin{tabular}{l|cc|cc|cc}
            \toprule
                                      & \multicolumn{2}{c|}{\textbf{Stairs}} & \multicolumn{2}{c|}{\textbf{Gaps}} & \multicolumn{2}{c}{\textbf{Platforms}}                                                                                   \\
            \textbf{Method}           & SR                                   & PDR$\downarrow$                    & SR                                     & PDR$\downarrow$          & SR                        & PDR$\downarrow$          \\
            \midrule
            Single Critic/Disc.       & 84.0$_{\pm 0.5}$                     & 19.2$_{\pm 0.3}$                   & 83.8$_{\pm 0.4}$                       & 19.2$_{\pm 0.3}$         & 76.3$_{\pm 0.6}$          & 25.4$_{\pm 0.4}$         \\
            Multi-Critic/Disc. (Ours) & \textbf{98.9}$_{\pm 0.4}$            & \textbf{5.7}$_{\pm 0.2}$           & \textbf{99.3}$_{\pm 0.2}$              & \textbf{4.5}$_{\pm 0.2}$ & \textbf{98.4}$_{\pm 0.5}$ & \textbf{7.2}$_{\pm 0.3}$ \\
            \midrule
            $\Delta$                  & +14.9                                & -13.5                              & +15.5                                  & -14.7                    & +22.1                     & -18.2                    \\
            \bottomrule
        \end{tabular}
    }
    \caption{Ablation study on multi-critic and multi-discriminator architecture. Stairs column reports the average of ascending (99.2\%) and descending (98.6\%) configurations. Results report mean$_{\pm \text{std}}$ over 5 random seeds.}
    \label{tab:ablation_critic}
    \vspace{-10pt}
\end{table}

\subsubsection{Distillation Objectives}

Table~\ref{tab:ablation_distillation} evaluates the distillation loss
components. The denoising objective $\mathcal{L}_{\text{denoise}}$ contributes
5.5\% SR improvement by enforcing consistent latent representations between
clean and augmented depth inputs. KL regularization $\mathcal{L}_{\text{kl}}$
provides 2.8\% SR improvement by preventing representation collapse. Using
behavior cloning alone achieves only 86.0\% SR with 17.7\% PDR, demonstrating
that auxiliary objectives are essential for robust vision-based deployment.

\begin{table}[h]
    \centering
    \begin{tabular}{lccc}
        \toprule
        \textbf{Configuration}             & \textbf{SR (\%)}             & \textbf{P ($\times 10^{1}$ W)}$\downarrow$ & \textbf{PDR (\%)}$\downarrow$ \\
        \midrule
        Full                               & \textbf{98.9}$_{\pm 0.4}$    & \textbf{27.7}$_{\pm 0.3}$                  & \textbf{5.8}$_{\pm 0.2}$      \\
        \midrule
        w/o $\mathcal{L}_{\text{denoise}}$ & 93.4$_{\pm 0.5}$             & 32.3$_{\pm 0.4}$                           & 13.3$_{\pm 0.3}$              \\
        w/o $\mathcal{L}_{\text{kl}}$      & \underline{96.1}$_{\pm 0.4}$ & \underline{29.5}$_{\pm 0.3}$               & \underline{8.9}$_{\pm 0.3}$   \\
        BC Only                            & 86.0$_{\pm 0.6}$             & 32.4$_{\pm 0.4}$                           & 17.7$_{\pm 0.4}$              \\
        \bottomrule
    \end{tabular}
    \caption{Ablation study on distillation loss components. Results report mean$_{\pm \text{std}}$ over 5 random seeds.}
    \label{tab:ablation_distillation}
    \vspace{-15pt}
\end{table}

\subsection{Latent Space Analysis}\label{sec:latent_analysis}

\begin{figure}[t]
    \centering
    \includegraphics[width=0.45\textwidth]{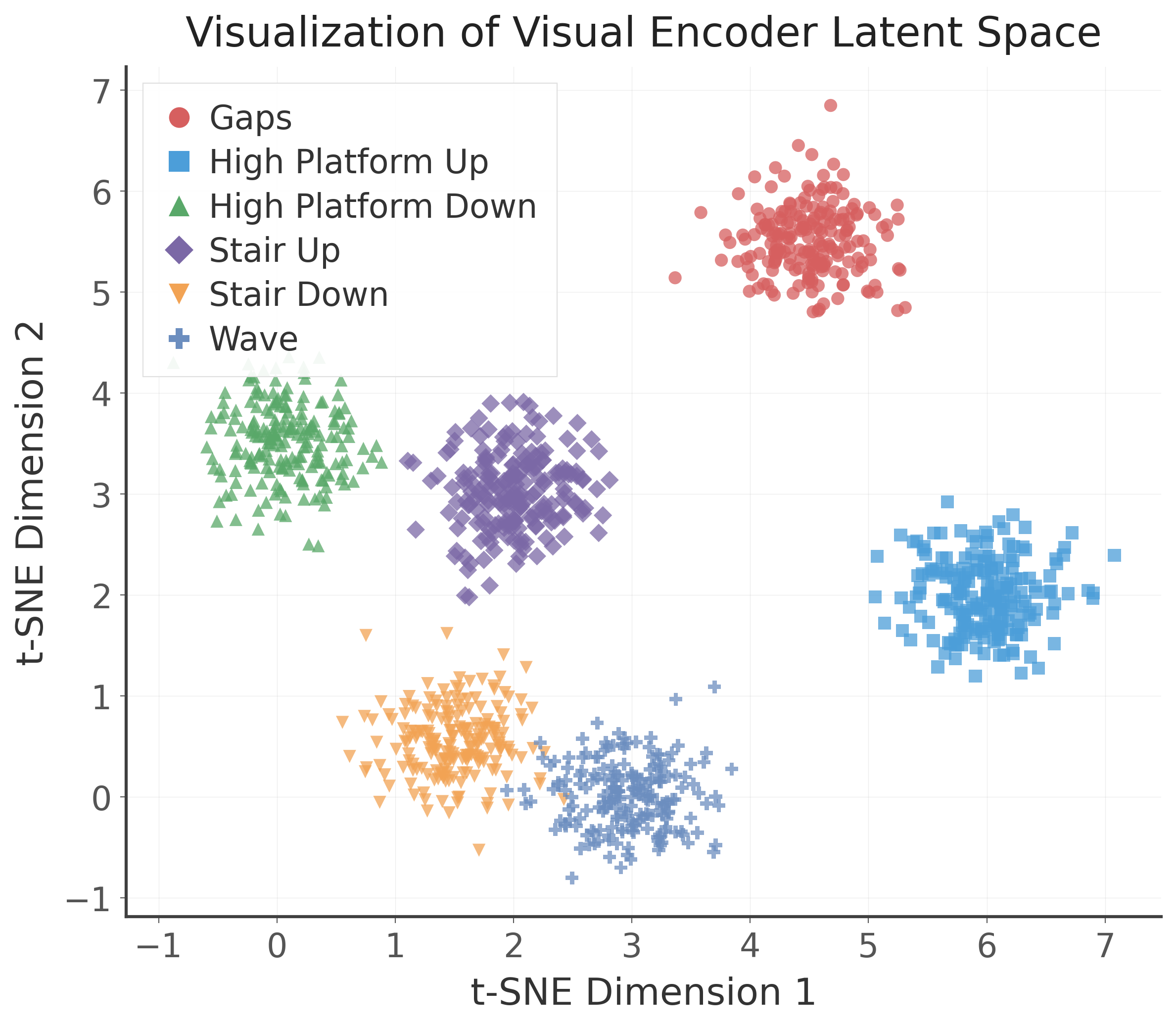}
    \caption{t-SNE visualization of the depth encoder's latent space across six terrain types. Each terrain forms a distinct cluster, demonstrating effective terrain-specific representation learning despite realistic sensor noise.}
    \label{fig:latent_vis}
    \vspace{-15pt}
\end{figure}

To understand how our depth encoder represents different terrains, we visualize
the learned latent space using t-SNE dimensionality reduction.
Figure~\ref{fig:latent_vis} shows the latent embeddings of augmented depth
observations across six terrain types.

The visualization reveals that each terrain type, stair ascending, stair
descending, gap, platform ascending, platform descending, and rough terrain,
forms a distinct, well-separated cluster. 
This clear separation indicates that our encoder
learns to extract terrain-specific geometric features despite realistic sensor
noise, providing the policy with discriminative information for appropriate
locomotion strategies.

\subsection{Real-World Deployment}\label{sec:real_world}

We validate our approach on a full-sized humanoid robot equipped with a stereo
depth camera. The policy runs entirely onboard at 50 Hz without any fine-tuning
from simulation training.

\subsubsection{Test Scenarios}

We evaluate four real-world scenarios: (1) outdoor stairs with 15 cm step
height (ascending and descending), (2) outdoor platforms with approximately 40
cm height, (3) extended staircases with 30+ consecutive steps, and (4) wide
gaps exceeding 45 cm width.

\subsubsection{Quantitative Results}

Table~\ref{tab:real_world} reports results from real-world deployment. The
policy achieves 97.8\% overall success rate (88/90 trials), with perfect
performance on five out of six test conditions.
The extended staircase test validates long-horizon stability over 30+
consecutive steps without accumulated drift.

\subsubsection{Failure Analysis}

The performance gap between stair ascending (100\%) and descending (86.7\%)
reveals an inherent asymmetry in bipedal locomotion: during descent,
gravitational acceleration amplifies minor errors into irreversible forward
momentum, whereas similar misjudgments during ascent remain recoverable.
Additionally, the target foot placement is partially occluded by the current
step edge, precisely where stereo matching artifacts are most pronounced. These
observations suggest that robust stair descent demands higher-resolution
near-field depth sensing or predictive foot placement strategies.

\begin{table}[h]
    \centering
    \begin{tabular}{lcc}
        \toprule
        \textbf{Scenario}       & \textbf{Success} & \textbf{Rate (\%)} \\
        \midrule
        Outdoor Stairs (up)     & 15/15            & 100.0              \\
        Outdoor Stairs (down)   & 13/15            & 86.7               \\
        Outdoor Platform (up)   & 15/15            & 100.0              \\
        Outdoor Platform (down) & 15/15            & 100.0              \\
        Extended Staircase      & 15/15            & 100.0              \\
        Wide Gap Crossing       & 15/15            & 100.0              \\
        \midrule
        \textbf{Overall}        & \textbf{88/90}   & \textbf{97.8}      \\
        \bottomrule
    \end{tabular}
    \caption{Real-world deployment success rates across 15 trials per scenario.}
    \label{tab:real_world}
    \vspace{-5pt}
\end{table}

\subsubsection{Cross-Platform Generalization}

To assess whether our training pipeline generalizes beyond the primary
platform, we apply the same methodology to a Unitree G1 humanoid robot equipped
with an Intel RealSense D435i depth camera, a different robot morphology and
sensor modality. Preliminary results are provided in
Appendix~\ref{append:cross_platform}, suggesting that our depth augmentation
and multi-critic framework can be effectively transferred to new hardware
configurations.
\section{Conclusion and Limitations}\label{sec:conclusion}

We presented a framework for vision-based humanoid locomotion that combines
realistic depth sensor simulation, multi-critic reinforcement learning, and
vision-aware behavior distillation. Our depth augmentation pipeline models
stereo fusion artifacts, depth-dependent noise, and calibration uncertainties,
while terrain-specific critics and discriminators enable unified control across
diverse scenarios. Experiments on RDT-Bench and real-world deployment
demonstrate strong performance across diverse terrains without fine-tuning.
However, stair descent remains less robust than ascent due to gravitational
amplification of minor errors and occlusion of target footholds by step edges,
which motivates future work on higher resolution near-field depth sensing and
predictive foot placement strategies.

\clearpage
\bibliographystyle{plainnat}
\bibliography{references}

\clearpage
\onecolumn

\begin{center}
    \LARGE\bfseries \appendix
\end{center}

\subsection{Network Architectures}\label{append:network_architectures}

We present the detailed network architectures for the teacher policy, student
policy, and multi-critic networks used in our framework. The teacher and
student policies share identical architectures except for their exteroceptive
encoders: the teacher uses an MLP to process privileged height scan
observations, while the student uses a CNN to process depth images. During
distillation, the student's CNN encoder learns to produce latent
representations that match the teacher's MLP encoder output, enabling seamless
transfer of the learned locomotion policy from privileged to vision-based
observations.

\begin{table}[h]
    \centering
    \small

    \begin{tabular}{lc}
        \toprule
        Component         & Configuration                           \\
        \midrule
        \multicolumn{2}{c}{\textit{Height Scan Encoder}}            \\
        \midrule
        Input Layer       & Height Scan (21 × 33 = 693)             \\
        Hidden Layer 1    & Linear(693 → 512) + SiLU                \\
        Hidden Layer 2    & Linear(512 → 256) + SiLU                \\
        Output Layer      & Linear(256 → 128)                       \\
        \midrule
        \multicolumn{2}{c}{\textit{Recurrent Module}}               \\
        \midrule
        Input             & Concat(Proprio[96], Encoder[128]) = 224 \\
        RNN Type          & GRU (1 layer)                           \\
        Hidden Dimension  & 256                                     \\
        \midrule
        \multicolumn{2}{c}{\textit{Actor MLP}}                      \\
        \midrule
        Input Layer       & GRU Output (256)                        \\
        Hidden Layer 1    & Linear(256 → 512) + SiLU                \\
        Hidden Layer 2    & Linear(512 → 256) + SiLU                \\
        Hidden Layer 3    & Linear(256 → 128) + SiLU                \\
        Output Layer      & Linear(128 → 29)                        \\
        \midrule
        \multicolumn{2}{c}{\textit{Policy Distribution}}            \\
        \midrule
        Distribution Type & Gaussian                                \\
        Initial Noise Std & 1.0                                     \\
        \bottomrule
    \end{tabular}
    \caption{Teacher Policy Network Architecture}
    \label{tab:teacher_arch}
\end{table}

\begin{table}[h]
    \centering
    \small

    \begin{tabular}{lc}
        \toprule
        Component          & Configuration                           \\
        \midrule
        \multicolumn{2}{c}{\textit{Depth Image Encoder (CNN)}}       \\
        \midrule
        Input Layer        & Depth Image (1 × 24 × 32)               \\
        Conv Layer 1       & Conv2d(1 → 32, k=3, s=2, p=1) + SiLU    \\
        Conv Layer 2       & Conv2d(32 → 64, k=3, s=2, p=1) + SiLU   \\
        Conv Layer 3       & Conv2d(64 → 128, k=3, s=2, p=1) + SiLU  \\
        Flatten            & 128 × 3 × 4 = 1536                      \\
        Output Layer       & Linear(1536 → 128)                      \\
        \midrule
        \multicolumn{2}{c}{\textit{Recurrent Module}}                \\
        \midrule
        Input              & Concat(Proprio[96], Encoder[128]) = 224 \\
        RNN Type           & GRU (1 layer)                           \\
        Hidden Dimension   & 256                                     \\
        \midrule
        \multicolumn{2}{c}{\textit{Actor MLP}}                       \\
        \midrule
        Input Layer        & GRU Output (256)                        \\
        Hidden Layer 1     & Linear(256 → 512) + SiLU                \\
        Hidden Layer 2     & Linear(512 → 256) + SiLU                \\
        Hidden Layer 3     & Linear(256 → 128) + SiLU                \\
        Output Layer       & Linear(128 → 29)                        \\
        \midrule
        \multicolumn{2}{c}{\textit{Policy Distribution}}             \\
        \midrule
        Distribution Type  & Gaussian                                \\
        Constant Noise Std & 0.1                                     \\
        \bottomrule
    \end{tabular}
    \caption{Student Policy Network Architecture}
    \label{tab:student_arch}
\end{table}

\begin{table}[h]
    \centering
    \small

    \begin{tabular}{lc}
        \toprule
        Component        & Configuration                           \\
        \midrule
        \multicolumn{2}{c}{\textit{Height Scan Encoder (Shared)}}  \\
        \midrule
        Input Layer      & Height Scan (21 × 33 = 693)             \\
        Hidden Layer 1   & Linear(693 → 512) + SiLU                \\
        Hidden Layer 2   & Linear(512 → 256) + SiLU                \\
        Output Layer     & Linear(256 → 128)                       \\
        \midrule
        \multicolumn{2}{c}{\textit{Recurrent Module (Shared)}}     \\
        \midrule
        Input            & Concat(Proprio[96], Encoder[128]) = 224 \\
        RNN Type         & GRU (1 layer)                           \\
        Hidden Dimension & 256                                     \\
        \midrule
        \multicolumn{2}{c}{\textit{Critic MLP (Shared Backbone)}}  \\
        \midrule
        Input Layer      & GRU Output (256)                        \\
        Hidden Layer 1   & Linear(256 → 512) + SiLU                \\
        Hidden Layer 2   & Linear(512 → 256) + SiLU                \\
        Hidden Layer 3   & Linear(256 → 128) + SiLU                \\
        \midrule
        \multicolumn{2}{c}{\textit{Terrain-Specific Output Heads}} \\
        \midrule
        Stair Head       & Linear(128 → 1)                         \\
        Gap Head         & Linear(128 → 1)                         \\
        General Head     & Linear(128 → 1)                         \\
        \bottomrule
    \end{tabular}
    \caption{Multi-Critic Network Architecture}
    \label{tab:critic_arch}
\end{table}


\subsection{Implementation Details}

We provide the detailed hyperparameters for our policy distillation framework
in Table~\ref{tab:distillation_config}. The distillation process transfers the
learned locomotion policy from the teacher (which operates on privileged height
scan observations) to the student (which operates on depth images). We employ a
cosine annealing learning rate schedule with warm-up to ensure stable training
dynamics. The loss coefficients are carefully balanced to ensure that behavior
cloning provides the primary learning signal while the denoising and KL
regularization objectives contribute to robustness against sensor noise and
prevent representation collapse, respectively. We use gradient accumulation to
stabilize training with large effective batch sizes and apply gradient clipping
to prevent training instabilities.

\begin{table}[ht]
    \centering
    \small
    \begin{tabular}{ll}
        \toprule
        \textbf{Parameter}                               & \textbf{Value}                                               \\
        \midrule
        \multicolumn{2}{l}{\textit{(a) Training Hyperparameters}}                                                       \\
        \cdashline{1-2}\noalign{\vskip 0.3mm}
        Steps per environment per iteration              & 800                                                          \\
        Total training iterations                        & 4000                                                         \\
        Learning rate schedule                           & One Cycle~\cite{smith2018superconvergencefasttrainingneural} \\
        Initial learning rate                            & $1 \times 10^{-3}$                                           \\
        Div factor (peak/init)                           & 10.0                                                         \\
        Final div factor (final/init)                    & 50.0                                                         \\
        Gradient accumulation steps                      & 10                                                           \\
        Max gradient norm                                & 1.0                                                          \\
        \midrule
        \multicolumn{2}{l}{\textit{(b) Loss Coefficients}}                                                              \\
        \cdashline{1-2}\noalign{\vskip 0.3mm}
        Behavior loss weight                             & $1.0$                                                        \\
        Denoising loss coeff. $\lambda_{\text{denoise}}$ & $0.1$                                                        \\
        KL loss coeff. $\lambda_{\text{kl}}$             & $0.1$                                                        \\
        EMA decay                                        & $0.997$                                                      \\
        \bottomrule
    \end{tabular}
    \caption{Policy Distillation Configuration}
    \label{tab:distillation_config}
    \vspace{-10pt}
\end{table}



\subsection{CycleGAN Hyperparameters}\label{append:cyclegan_hyperparameters}

We train a CycleGAN model to translate between simulated and real-world depth
images for evaluation purposes. The real-world dataset consists of
approximately 2 hours of depth recordings from the humanoid robot traversing
various terrains, yielding approximately 200,000 depth frames. The CycleGAN is
used \textit{exclusively for evaluation} to inject realistic depth artifacts
into simulation, enabling fair comparison of sim-to-real transfer capabilities
across different training augmentation strategies.

\subsubsection{Translation Quality Evaluation}

To validate that our CycleGAN produces realistic depth translations, we
evaluate the model using standard image-to-image translation metrics on a
held-out test set comprising 10\% of the data (approximately 20,000 frames).

\begin{table}[h]
    \centering
    \small
    \begin{tabular}{lcc}
        \toprule
        \textbf{Metric}                            & \textbf{Sim$\rightarrow$Real} & \textbf{Real$\rightarrow$Sim} \\
        \midrule
        \multicolumn{3}{l}{\textit{Translation Quality}}                                                           \\
        \cdashline{1-3}\noalign{\vskip 0.3mm}
        FID $\downarrow$ (no translation baseline) & \multicolumn{2}{c}{67.2}                                      \\
        FID $\downarrow$ (after CycleGAN)          & 23.4                          & 21.7                          \\
        KID ($\times 10^{-3}$) $\downarrow$        & 8.2                           & 7.6                           \\
        \midrule
        \multicolumn{3}{l}{\textit{Cycle Consistency}}                                                             \\
        \cdashline{1-3}\noalign{\vskip 0.3mm}
        SSIM $\uparrow$                            & 0.89                          & 0.91                          \\
        PSNR $\uparrow$                            & 28.3 dB                       & 29.1 dB                       \\
        LPIPS $\downarrow$                         & 0.12                          & 0.11                          \\
        \bottomrule
    \end{tabular}
    \caption{CycleGAN Translation Quality Metrics. \textbf{Translation Quality}: FID and KID measure distributional distance between translated images $G(x_{\text{source}})$ and real target domain images $x_{\text{target}}$. The baseline FID (67.2) is computed between raw simulation and real depth images without translation. \textbf{Cycle Consistency}: SSIM, PSNR, and LPIPS evaluate reconstruction quality of $F(G(x)) \approx x$, measuring how well geometric structure is preserved through the translation cycle. Images are resized to $128 \times 128$ before FID computation using bilinear interpolation.}
    \label{tab:cyclegan_quality}
    \vspace{-10pt}
\end{table}

The FID score of 23.4 for Sim$\rightarrow$Real translation indicates that the
generated depth images closely match the statistical properties of real sensor
outputs. The high cycle-consistency metrics (SSIM = 0.89, PSNR = 28.3 dB)
confirm that the translation preserves geometric structure while introducing
realistic sensor artifacts.

\begin{table}[h]
    \centering
    \small

    \begin{tabular}{ll}
        \toprule
        \textbf{Parameter}                            & \textbf{Value}            \\
        \midrule
        \multicolumn{2}{l}{\textit{(a) Dataset Statistics}}                       \\
        \cdashline{1-2}\noalign{\vskip 0.3mm}
        Total real-world recording time               & $\sim$2 hours             \\
        Total depth frames                            & $\sim$200,000             \\
        Frame resolution                              & 24 $\times$ 32            \\
        Train/Val split                               & 90\% / 10\%               \\
        \midrule
        \multicolumn{2}{l}{\textit{(b) Network Architecture}}                     \\
        \cdashline{1-2}\noalign{\vskip 0.3mm}
        Generator architecture                        & ResNet (9 blocks)         \\
        Discriminator architecture                    & PatchGAN (70 $\times$ 70) \\
        Number of filters (first layer)               & 64                        \\
        Normalization                                 & Instance Normalization    \\
        \midrule
        \multicolumn{2}{l}{\textit{(c) Training Hyperparameters}}                 \\
        \cdashline{1-2}\noalign{\vskip 0.3mm}
        Batch size                                    & 64                        \\
        Total epochs                                  & 200                       \\
        Optimizer                                     & Adam                      \\
        Learning rate                                 & $2 \times 10^{-4}$        \\
        $\beta_1$, $\beta_2$                          & 0.5, 0.999                \\
        Learning rate decay                           & Linear (after epoch 100)  \\
        \midrule
        \multicolumn{2}{l}{\textit{(d) Loss Weights}}                             \\
        \cdashline{1-2}\noalign{\vskip 0.3mm}
        Adversarial loss weight                       & 1.0                       \\
        Cycle consistency loss weight $\lambda_{cyc}$ & 10.0                      \\
        Identity loss weight $\lambda_{idt}$          & 0.5                       \\
        \bottomrule
    \end{tabular}
    \caption{CycleGAN Training Configuration}
    \label{tab:cyclegan_config}
    \vspace{-10pt}
\end{table}

\subsection{Terrain Specific Rewards}\label{append:terrain_spec_rewards}

Our multi-critic framework employs terrain-specific reward functions to
encourage appropriate locomotion strategies for different obstacle types. We
detail the key terrain-specific rewards below.

\subsubsection{Velocity Tracking Rewards}

We employ two distinct velocity tracking formulations depending on terrain
requirements:

\textbf{Exponential Velocity Tracking} ($r_{\text{vel}}^{\text{exp}}$) is used for
stairs, platforms, and rough terrain, where precise velocity regulation ensures
stable foot placement:
\begin{equation}
    r_{\text{vel}}^{\text{exp}} = \exp\left(-\frac{\|\mathbf{v}_{xy}^{\text{cmd}} - \mathbf{v}_{xy}^{\text{robot}}\|^2}{\sigma^2}\right)
\end{equation}
where $\mathbf{v}_{xy}^{\text{cmd}}$ denotes the commanded velocity,
$\mathbf{v}_{xy}^{\text{robot}}$ denotes the robot's base linear velocity in the
body frame, and $\sigma$ is a temperature parameter controlling tracking
precision. This exponential kernel provides smooth gradients that encourage
accurate velocity following, which is critical for controlled stepping on
elevated surfaces.

\textbf{Directional Velocity Tracking} ($r_{\text{vel}}^{\text{dir}}$) is used
for gap terrain, where the robot benefits from moving faster than the commanded
velocity to execute dynamic crossing motions:
\begin{equation}
    r_{\text{vel}}^{\text{dir}} = \frac{\min\left(\mathbf{v}_{xy}^{\text{robot}} \cdot \hat{\mathbf{d}}_{\text{cmd}},\, \|\mathbf{v}_{xy}^{\text{cmd}}\|\right)}{\|\mathbf{v}_{xy}^{\text{cmd}}\| + \epsilon}
\end{equation}
where $\hat{\mathbf{d}}_{\text{cmd}} = \mathbf{v}_{xy}^{\text{cmd}} / \|\mathbf{v}_{xy}^{\text{cmd}}\|$
is the unit direction of the commanded velocity, and $\epsilon$ is a small
constant for numerical stability. This formulation rewards movement in the
commanded direction without penalizing speeds exceeding the command magnitude,
enabling the robot to build momentum for successful gap traversal.

\subsubsection{Feet Contact Height Reward}

For stairs and platform terrains, we introduce a feet contact height reward
($r_{\text{contact}}$) that encourages the robot to place its feet on flat,
stable surfaces:
\begin{equation}
    r_{\text{contact}} = \sum_{f \in \{\text{left}, \text{right}\}} \mathbb{1}_{\text{contact}}^f \cdot \text{std}\left(\text{clip}(h_f^{\text{scan}}, -h_{\max}, h_{\max})\right)
\end{equation}
where $h_f^{\text{scan}}$ denotes the height scan measurements around foot $f$
relative to the foot position, $\mathbb{1}_{\text{contact}}^f$ is an indicator
function that equals 1 when foot $f$ is in contact with the ground, and
$h_{\max}$ is a clipping threshold. The standard deviation of clipped height
values around each foot serves as a measure of surface irregularity, lower
values indicate flatter contact surfaces. This reward is applied as a penalty
(with negative weight) to discourage foot placement on edges or uneven surfaces,
which is particularly important for stair and platform traversal where precise
foot placement determines stability.

This reward is \textit{not} applied to gap and rough terrains: gap crossing
requires dynamic leaping motions where contact surface analysis is less
relevant, while rough terrain inherently features irregular surfaces where
penalizing height variation would be counterproductive.

\begin{table}[h]
    \centering
    \small

    \begin{tabular}{lccc}
        \toprule
        \textbf{Reward}               & \textbf{Stairs/Platforms} & \textbf{Gaps} & \textbf{Rough} \\
        \midrule
        $r_{\text{vel}}^{\text{exp}}$ & \checkmark                & --            & \checkmark     \\
        $r_{\text{vel}}^{\text{dir}}$ & --                        & \checkmark    & --             \\
        $r_{\text{contact}}$          & \checkmark                & --            & --             \\
        \bottomrule
    \end{tabular}
    \caption{Terrain-Specific Reward Configuration}
    \label{tab:terrain_rewards}
    \vspace{-10pt}
\end{table}

\subsection{Cross-Platform Validation}\label{append:cross_platform}

We conduct a preliminary cross-platform validation to assess whether the
learned policy generalizes beyond the primary training platform. We deploy our
policy on a Unitree G1 humanoid robot, which differs from our primary platform
in both kinematic structure and sensor configuration (Intel RealSense D435i vs.
Orbbec Gemini 336L).

\begin{table}[h]
    \centering
    \small

    \begin{tabular}{lcc}
        \toprule
        \textbf{Scenario} & \textbf{Success} & \textbf{Rate (\%)} \\
        \midrule
        Stair Ascending   & 15/15            & 100.0              \\
        \bottomrule
    \end{tabular}
    \caption{Cross-Platform Validation on Unitree G1}
    \label{tab:cross_platform}
\end{table}

\begin{figure}[h]
    \centering
    \includegraphics[width=1.0\textwidth]{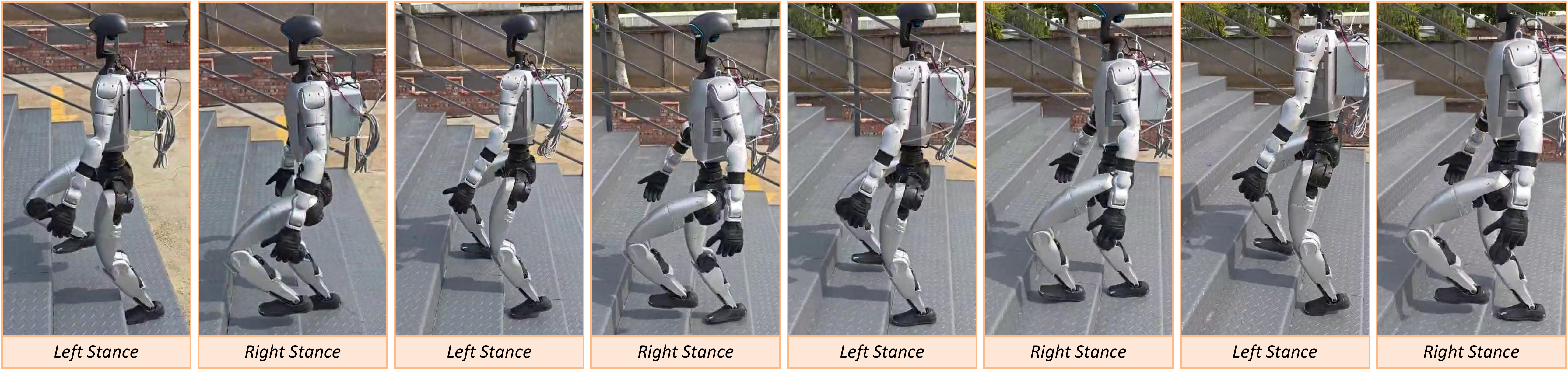}
    \caption{Cross-platform deployment on Unitree G1 humanoid robot ascending
        outdoor stairs. The policy transfers zero-shot from training on a different
        platform with different depth sensor, demonstrating the generality of the
        learned depth representations.}
    \label{fig:g1_up_stair}
    \vspace{-15pt}
\end{figure}

As shown in Table~\ref{tab:cross_platform} and Figure~\ref{fig:g1_up_stair},
the policy achieves perfect success rate on the stair ascending task without
any platform-specific adaptation. This result provides initial evidence that
our depth augmentation strategy learns transferable geometric representations
rather than sensor-specific artifacts. However, we note that this validation is
limited to a single terrain type; comprehensive cross-platform benchmarking
across gaps, platforms, and descending scenarios remains an important direction
for future work.

\subsection{Additional Augmentation Results}\label{append:additional_aug_results}
\begin{figure}[h]
    \centering
    \includegraphics[width=0.8\textwidth]{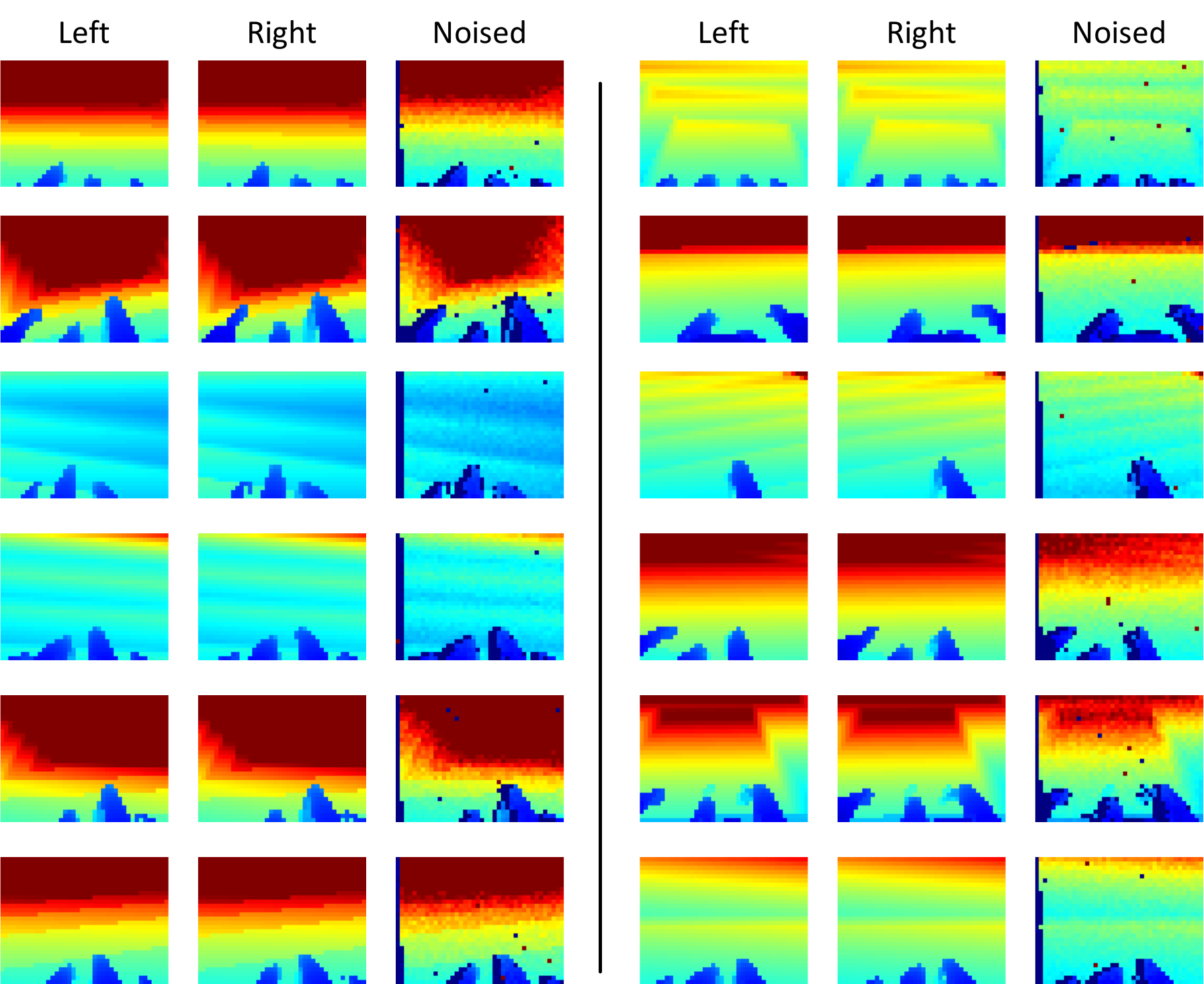}
    \caption{Additional depth augmentation examples across diverse terrains.
        Each triplet shows (left to right): left camera depth, right camera depth,
        and augmented output before spatial cropping. Depth values are normalized to
        $[0, 2]$ m and rendered as color maps (cool = near, warm = far). The
        augmented images exhibit realistic stereo fusion holes (black regions),
        depth-dependent noise, and structured Perlin patterns while preserving
        terrain geometry essential for locomotion control.}
    \label{fig:many_aug}
    \vspace{-15pt}
\end{figure}

We provide additional visualization examples of our depth augmentation pipeline
in Figure~\ref{fig:many_aug}. Each row shows a triplet consisting of: (1) the
left camera depth image, (2) the right camera depth image, and (3) the
augmented depth output before spatial cropping. All depth values are normalized
to the range $[0, 2]$ meters and converted to color maps for intuitive
visualization, where cooler colors (blue/purple) indicate closer surfaces and
warmer colors (red/yellow) represent farther distances.

\end{document}